\pdfoutput=1

\documentclass[11pt]{article}

\usepackage[preprint]{acl}

\usepackage{times}
\usepackage{latexsym}

\usepackage[T1]{fontenc}

\usepackage[utf8]{inputenc}

\usepackage{microtype}

\usepackage{inconsolata}

\usepackage{graphicx}

\usepackage{amsfonts}

\definecolor{mylightgray}{RGB}{220,220,220}
\definecolor{mylightblue}{RGB}{202, 241, 202}
\newcommand{\modelname}{\texttt{Direct-Inverse Discriminative Prompting}} 
\newcommand{\modelnameshort}{\texttt{Direct-Inverse Prompting}}  
\newcommand{\direct}{\texttt{Direct Prompt}}  
\newcommand{\inverse}{\texttt{Inverse Prompt}}  
\newcommand{\hybrid}{\texttt{Combination}}  

\title{\modelnameshort: Analyzing LLMs' Discriminative Capacity \\in Self-Improving Generation}

\author{
  Jihyun Janice Ahn\textsuperscript{\rm $\spadesuit$} \quad
   Ryo Kamoi\textsuperscript{\rm $\spadesuit$} \quad
  Lu Cheng\textsuperscript{\rm $\diamondsuit$} \quad
  \\
  \textbf{Rui Zhang}\textsuperscript{\rm $\spadesuit$} \quad
  \and
  \textbf{Wenpeng Yin}\textsuperscript{\rm $\spadesuit$}
  \\
  \textsuperscript{\rm $\spadesuit$}The Pennsylvania State University
  \ 
  \textsuperscript{\rm $\diamondsuit$} University of Illinois at Chicago
  \\
  {\texttt{\{jfa5672, wenpeng\}@psu.edu};}
  \ 
  { \texttt{lucheng@uic.edu}}
}

\begin{document}
\maketitle
\begin{abstract}
Mainstream LLM research has primarily focused on enhancing their generative capabilities. However, even the most advanced LLMs experience uncertainty in their outputs, often producing varied results on different runs or when faced with minor changes in input, despite no substantial change in content. Given multiple responses from the same LLM to the same input, we advocate leveraging the LLMs' discriminative capability to reduce this generative uncertainty, aiding in identifying the correct answers. Specifically, we propose and analyze three discriminative prompts: \direct, \inverse, and \hybrid, to explore the potential of both closed-source and open-source LLMs in self-improving their generative performance on two benchmark datasets. Our insights reveal which discriminative prompt is most promising and when to use it. To our knowledge, this is the first work to systematically analyze LLMs' discriminative capacity to address generative uncertainty.
\end{abstract}

\section{Introduction}

Generative AI is revolutionizing various fields by utilizing large language models (LLMs) trained to generate human-like responses based on given instructions. Despite the increasing strength of existing LLMs in terms of generation capability, a widely recognized issue is their uncertainty in responses to inputs—the same model may produce significantly different responses on different runs or to equivalently varied inputs.

Previous studies have explored LLMs' self-improving capability that either relied on external human/tool supervision \cite{DBLP:journals/corr/abs-2308-04592,DBLP:conf/eacl/PaulIPBBWF24,DBLP:journals/corr/abs-2305-11738,DBLP:journals/corr/abs-2304-05128,DBLP:journals/corr/abs-2306-09896,DBLP:conf/acl/GaoDPCCFZLLJG23} or have not successfully explored the inner capabilities of LLMs, such as their own discriminative capability, to reduce uncertainty \cite{jiang2024selfincorrect}. 
We argue that LLMs should focus on both their generative and discriminative capabilities. In this work, we explore various discriminative capabilities of LLMs to reduce the uncertainty of their self-improving generations.

Specifically, we propose and analyze three types of discriminative prompts to identify the most promising answer from a group of generated responses: \direct: directly asking the LLM which responses are correct; \inverse: contrasting \direct by asking which responses are incorrect; \hybrid: combining \direct and \inverse, since intuitively they perform the same reasoning process from complementary perspectives.

We conduct analyses with two closed-source LLMs (GPT-4 \cite{openai2023gpt4} and GPT-4o \cite{OpenAI2024GPT4o}) and two open-source LLMs (Llama-3-8B-Instruct \cite{Meta2024Llama3} and MetaMath-7B-V1.0 \cite{yu2023metamath}) on two math-related datasets, MATH \cite{DBLPHendrycksBKABTS21} and MathQA \cite{DBLPAminiGLKCH19}. We observe: i) For closed-source LLMs, using discriminative capability, either \direct~or \inverse, is highly effective for reducing uncertainty in self-improving generations. ii) For open-source LLMs, if not instruction-tuned, using discriminative capability is not recommended. Even if instruction-tuned, only \direct~is recommended due to likely issues with understanding negation in \inverse.

Our contributions are threefold:
\begin{itemize}
    \item Proposing \modelname, a multi-angle complementary method, to assess LLMs' discriminative capability in self-improving generation;
    \item The first systematic analysis of the potential of LLMs' discriminative capability to reduce generative uncertainty;
    \item Providing insights and suggestions for future users on how to effectively utilize LLMs' discriminative capability in practice.
\end{itemize}

\section{Related Work}

\paragraph{LLM self-improves generation.} Various methods are being devised to increase the certainty of LLM-generated answers.  Chain-of-Thought \cite{wei2023chainofthought} tries to add a detailed reasoning path from the input to the output answer so that the answer is more explainable and certain.  Self-Consistency \cite{wang2023selfconsistency}   has the LLM solve the same problem multiple times to obtain several results. A majority vote is then conducted to choose the most consistent result as the final answer. This approach guarantees a higher success rate than Chain-of-Thought. Based on this, diverse variants of Self-Consistency exist; for example, Universal Self-Consistency \cite{chen2023universal}, which includes reasoning to select the most consistent value as the final answer, or Early Stop Self-Consistency \cite{li2024escape}, which reduces the number of answer sets used in the majority vote to save cost and time. It is worth mentioning that the above approaches are fully unsupervised, namely no human or external signals are needed.

\paragraph{Exploring LLM discriminative capability to enhance generation.}

To assess the generative and discriminative capabilities of LLMs, \citet{DBLP:journals/corr/abs-2311-09184} and \citet{DBLP:journals/corr/abs-2305-17077} carried out experiments on summarization and planning problem, respectively.
The most related work, \cite{jiang2024selfincorrect}, concluded that LLMs struggle to enhance their generation performance through discriminative capability because their discriminative capability is not stronger than their generative capability. Our work differs from this study in two key ways: i) \citet{jiang2024selfincorrect} only considered a simplified discriminative prompt similar to our \direct. They provided the discriminative prompt with all the generated final answers without the reasoning paths. In contrast, our \direct~includes reasoning-path equipped answers, which we believe can help LLMs better determine the correct answer. ii) We further analyze another complementary discriminative capability expressed by \inverse. While \inverse~should theoretically yield the same conclusions if applied to humans, the inconsistency between \direct~and \inverse~in LLMs allows us to better understand their discriminative potential in reducing generative uncertainty. iii) Our findings suggest a different conclusion: LLMs' discriminative capabilities can indeed enhance their generation if used skillfully.


\section{\modelname}

Given multiple answer options by LLMs' generative process (here uses five for example), this section introduces our discriminative approach \modelname, that asks LLMs with \direct, \inverse, and finally combines their lens to find the most certain answer in self-improving generation.

\paragraph{\direct.} Here, we directly ask LLMs which options are correct with the following prompt:
\begin{center}
\colorbox{mylightblue}{\parbox{0.9\linewidth}{This problem [\emph{problem description}] has the following reasoning paths you generated:  `` A: [$path_1$]'',  ``B: [$path_2$]'',  ``C: [$path_3$]'', ``D: [$path_4$]'',  ``E: [$path_5$]''. Please output the correct ones.} }
\end{center}

\paragraph{\inverse.} Here, we ask LLMs which options are incorrect with the following prompt:
\begin{center}
\colorbox{mylightblue}{\parbox{0.9\linewidth}{This problem [\emph{problem description}] has the following reasoning paths you generated:  `` A: [$path_1$]'',  ``B: [$path_2$]'',  ``C: [$path_3$]'', ``D: [$path_4$]'',  ``E: [$path_5$]''. Please output the incorrect ones.} }
\end{center}

\paragraph{\hybrid.} As humans, when asked using both \direct~and \inverse~prompts, their answers should be consistent. However, this is not the case with LLMs, as our analysis in Section \ref{sec:analysis} shows. For instance, using \direct, an LLM may believe ``A and B'' are correct, but when asked using \inverse, it might believe ``B and C'' are incorrect, implying that ``A, D, and E'' are correct. \direct~and \inverse~reflect LLMs' discriminative analysis of the problem from different perspectives, and we combine their results to improve accuracy. Specifically, we run \direct~and \inverse~separately multiple times and select the final answer by identifying the most consensus among the responses.

\begin{table*}[t]
    \setlength{\tabcolsep}{5.5pt}
    \centering
    \begin{tabular}{l|rrrr|rrrr}
         & \multicolumn{4}{c|}{MATH} & \multicolumn{4}{c}{MathQA} \\
         &\multicolumn{1}{c}{GPT4} & \multicolumn{1}{c}{GPT-4o} & \multicolumn{1}{c}{Llama3} & \multicolumn{1}{c|}{MetaMath} &\multicolumn{1}{c}{GPT4} & \multicolumn{1}{c}{GPT-4o} & \multicolumn{1}{c}{Llama3} & \multicolumn{1}{c}{MetaMath} \\\hline
        Chain-of-Thought & 47.58 & 50.67 & 21.55 & 10.83 & 72.57 & 82.73 & 39.03 & 11.96\\
        Uni. Self-Consist. & 55.14 & 54.72 & 26.72 & 12.04 & 79.50 & 85.33 & 42.58 & 11.79\\
        \direct & 54.18 & \underline{\textbf{57.44}} & \underline{\textbf{27.54}} & 0.18 & \underline{81.64} & \underline{\textbf{86.73}} & \underline{\textbf{46.40} }& 0.00\\
        
        \inverse & 54.62 & \underline{55.48} & 18.08 & 0.06 & \underline{\textbf{82.34}} & \underline{86.40} & 37.45 & 0.00\\
        
        \hybrid & \underline{\textbf{56.44}} & \underline{56.82} & 25.98 & 0.24 & \underline{82.04} & \underline{86.63} & \underline{42.98} & 0.00
    \end{tabular}
    \caption{Comparing discriminative prompts \direct, \inverse, and \hybrid~on  LLMs. Bold: top score. Underline: surpass the Universal Self-Consistency.}
    \label{tab:mainresults}
\end{table*}

\section{Experiments}

\paragraph{Datasets.} Two datasets. An example of each dataset is given in appendix \ref{sec:appendix}.

\textbullet\enspace\texttt{MATH} \cite{DBLPHendrycksBKABTS21}: This dataset contains 7 types of open-ended math problems, including algebra and geometry, with average high school difficulty. For this project, we selected the entire test dataset of 5,000 problems. Each problem includes a ``problem'' label, representing the math word problem, and a ``solution'' label, which provides the explanation of how to solve the problem, including an answer formatted as \texttt{\$\textbackslash boxed\{$A$\}\$}, where $A$ is the answer. To maintain consistency, all models were instructed to return the final answer in the same format as the dataset.

\textbullet\enspace\texttt{MathQA} \cite{DBLPAminiGLKCH19}: This dataset includes 6 types of math problems, with college-level difficulty. We selected all 2,985 problems from the test dataset. Each entry in MathQA contains a ``problem,'' a ``rationale'' explaining how to solve it, ``options'' that list possible answers, and ``correct,'' indicating the correct answer from the options. LLMs were instructed to return the correct option's alphabet from the given choices.

\paragraph{LLMs.} i) Two closed-source LLMs: GPT-4 \cite{openai2023gpt4} and GPT-4o \cite{OpenAI2024GPT4o}. Both by OpenAI APIs. We do not consider more closed-source LLMs due to budget limits, and GPT-4 and GPT-4o are already widely recognized as the strongest LLMs. ii) Open-source LLMs: Llama-3 \cite{Meta2024Llama3} and MetaMath \cite{yu2023metamath}--a LLM specifically optimized for math problem solving. In our experiments, five A100 GPUs were used for running Llama-3 and MetaMath inference.

\paragraph{Baselines.} i) Chain-of-Thought (CoT) \cite{DBLP:conf/nips/Wei0SBIXCLZ22}. We run it three times and report the average performance. ii) Universal Self-Consistency \cite{chen2023universal}, the state-of-the-art approach CoT reasoning process five times, and finally choosing the answer with majority voting.

\paragraph{Setting.} To prevent the LLMs' responses to options like ``A, B, C, etc.'' from being biased due to their pretraining, we will shuffle these options and re-index them for each run. The final performance will be the average of three runs.

\section{Results}

\subsection{Main Results}
Table \ref{tab:mainresults} presents the main results comparing different discriminative prompts (\direct, \inverse, and \hybrid) of LLMs on the MATH and MathQA datasets. Here are some key observations:
\begin{itemize}
\setlength\itemsep{0.5em}
    \item  Discriminative prompts (\direct, \inverse, and \hybrid) do not work for MetaMath. This is because MetaMath was specifically optimized for solving math problems rather than following instructions. In our experiments, MetaMath responded to our discriminative prompts with noise and unstructured outputs, making answer parsing impossible.
    \item  Excluding MetaMath, \inverse~outperforms \direct~in 2 out of 6 cases, performs equally in one case (GPT-4o on MathQA), and underperforms in the remaining three cases. This is expected because negation is often more challenging for AI models to understand.
    \item  In most cases (except for MetaMath), both \direct~and \hybrid~outperform Universal Self-Consistency (and even \inverse~generally surpasses it on closed-source LLMs), indicating the effectiveness of using LLMs' discriminative capabilities to find the most certain answer.
\end{itemize}



\begin{table}[!t]
    \centering
    \begin{tabular}{l|r|r}
         & MATH & MathQA\\\hline
       GPT-4  & 36.88 &  23.75\\  
       GPT-4o  &  46.00 & 23.85 \\   
       Llama-3 &  97.34 & 97.96 \\
       MetaMath &  100.00 &  100.00\\
    \end{tabular}
    \caption{Conflicting percentage per dataset. }
    \label{tab:conflicts}
\end{table}

\begin{table}[!t]
    \centering
    \begin{tabular}{l|c|c}
         & MATH  & MathQA \\\hline
       GPT-4  &   71.86 / 25.49   &  89.02 / 58.81 \\  
       GPT-4o  & 77.93 / 30.30   &  93.36 / 62.64  \\   
       Llama-3 &  76.69 / 23.07   &  73.77 / 42.23 \\
       MetaMath &  0.00  / 0.12  &   0.00 / 0.00 \\
    \end{tabular}
    \caption{Fine-grained \hybrid~performance on agreed/disagreed responses of \direct~and \inverse~. 
    }
    \label{tab:finegrained}
\end{table}

\subsection{Analysis}\label{sec:analysis}

\paragraph{$\mathcal{Q}_1$: How frequently do LLMs experience uncertainty in their decisions, indicated by conflicts between \direct and \inverse?}
When \inverse~outputs, for instance, ``B, C'' as incorrect answers, we consider the remaining options, i.e., ``A, D, E'' as the correct answer inferred by \inverse. Conflicts arise when \direct~and \inverse~reach different conclusions. The conflict degree is calculated as the number of conflicts divided by the total number of problems for each dataset.

Table \ref{tab:conflicts} provides a summary of the severity of self-conflict within each LLM. GPT-4 demonstrates the highest consistency and self-confidence, with the lowest conflict percentages across both datasets. GPT-4o shows moderate consistency, performing better on the MathQA dataset than on MATH. Llama-3 exhibits the weakest performance in terms of consistency on the MathQA dataset, with the second-highest conflict rates in the MATH dataset, indicating its unreliability in this analysis. Lastly, MetaMath shows the highest conflict rates in both datasets having 100\% of conflict rates. These results underscore the enhanced reliability of advanced models like GPT-4. They also emphasize the interestingness of our work, which leverages the inconsistency in discriminative capability to enhance the certainty in generative 

\paragraph{$\mathcal{Q}_2$: How are LLMs performing when their choice is agreed or disagreed by \direct~and \inverse?} To answer this question, we check the fine-grained \hybrid~performance for the agreed and disagreed subsets between \direct~and \inverse.

Table \ref{tab:finegrained} presents the performance of LLMs when they are certain (both \direct~and \inverse~agree) or uncertain (they conflict).
It is clear that when \direct~and \inverse~agree, the answers are more likely to be correct, demonstrating significantly higher performance than both their disagreed subset and the overall dataset in Table \ref{tab:mainresults}. This further suggests that combining \direct~and \inverse~is an effective method for reducing uncertainty. 
If \direct~and \inverse~disagree, a comparison between Table \ref{tab:mainresults} and Table \ref{tab:finegrained} indicates that \direct~is the preferred approach. These conclusions generally apply to most LLMs, except for MetaMath, which is non-functional due to its pretraining limitations.


\paragraph{$\mathcal{Q}_3:$ When to suggest using \direct~and \inverse~to self-improve generation?} Based on Table \ref{tab:mainresults}, we can summarize two criteria: i) For top-performing closed-source LLMs like GPT-4 and GPT-4o, using either \direct~or \inverse, or their combination \hybrid, shows promise. These top LLMs perform similarly when \direct~and \inverse~are used separately. Combining them can result in robust performance, but the additional time and budget required for \hybrid~may not be appealing. Therefore, the concise conclusion for the top-performing closed-source models is that either \direct~or \inverse~is sufficient. ii) For open-source LLMs, the decision to try discriminative prompts depends on two factors: a) If the LLMs are not optimized to follow instructions, such as MetaMath, neither \direct~nor \inverse~is recommended. b) Even if the model is instruction-tuned, open-source LLMs are more likely to struggle with understanding negation, so only \direct~is strongly and exclusively recommended.

\section{Conclusion}

This study analyzed the development of LLM's discriminative capability to enhance self-improving generation performance. Specifically, we introduce \modelname, a multi-faceted complementary approach to evaluating LLMs' discriminative potential. Our findings indicate that both \direct~and \inverse~are effective for closed-source LLMs, while for open-source LLMs, using \direct~is highly and solely recommended.

\section*{Limitations}

Our study is limited by the fact that experiments were conducted using only two datasets. In addition, if budget permits, exploring more closed-source LLMs is preferred.

\section*{Ethics Statement}
This study uses publicly and automatically accessed datasets, and no ethical issues are present.
\bibliography{custom}

\appendix

\section{Example Appendix}
\label{sec:appendix}
\subsection{MATH}
\vspace{2mm}
\noindent
\fbox{
     \begin{minipage}{0.95\linewidth}
     $\mathcal{Q}$: \texttt{What is the 100th term of the arithmetic sequence 6, 10, 14, 18, ...?} \\
     $\mathcal{R}$: \texttt{The common difference is \$$10 - 6 = 4$\$, so the 100th term is \$$6+99\cdot 4=boxed\{402\}$\$.} 
     \end{minipage}
     }
\vspace{1mm}

where “$\mathcal{Q}$” denotes questions and “$\mathcal{R}$” for rationale. “$\mathcal{R}$” includes the answer in a specific format which is boxed\{\textit{A}\}, where \textit{A} is the answer for the problem.


\subsection{MathQA}
\vspace{2mm}
\noindent
\fbox{%
         \begin{minipage}{0.95\linewidth}
         $\mathcal{Q}$: \texttt{what will be the difference between simple and compound interest at 14 \% per annum on a sum of rs . 1000 after 4 years ?} \\
         $\mathcal{R}$: \texttt{s . i . = ( 1000 * 14 * 4 ) / 100 = rs . 560 c . i . = [ 1000 * ( 1 + 14 / 100 ) 4 - 1000 ] = rs . 689 difference = ( 689 - 560 ) = rs . 129 answer : a} \\
         $\mathcal{O}$: \texttt{a) 129 , b) 130 , c) 124 , d) 133 , e) 145}\\
         $\mathcal{A}$: \texttt{a}\\
         \end{minipage}
         }
\vspace{1mm}

where "$\mathcal{Q}$" denotes questions, "$\mathcal{R}$" for rationale, "$\mathcal{O}$" for options, and "$\mathcal{A}$" for answers.


\end{document}